# Artificial Dendritic Computation: The case for dendrites in neuromorphic circuits.

D. J. Mannion[1*], A. J. Kenyon[1]

[1]*Department of Electronic & Electrical Engineering, University College London (UCL), London, United Kingdom.*

*Corresponding author: daniel.mannion.13@ucl.ac.uk*

## Abstract

Bioinspired computing has focused on neurons and synapses with great success. However, the connections between these, the dendrites, also play an important role. In this paper, we investigate the motivation for replicating dendritic computation and present a framework to guide future attempts in their construction. The framework identifies key properties of the dendrites and presents an example of dendritic computation in the task of sound localisation. We evaluate the impact of dendrites on an BiLSTM neural network's performance, finding that dendritic pre-processing reduces the size of network required for a threshold performance.

## 1 Introduction & Background Theory

Neural networks consisting of neurons and synapses have proved successful in advancing the field of machine learning. The generalisability of the architecture has meant the fundamental concept of decision-making neurons and tuneable synaptic weights has been applied in a range of applications. But while the performance of deep neural networks has progressed there is still a large discrepancy between the power efficiency of deep neural networks and their biological counterparts [1]. Why does this discrepancy exist? Identifying differences between the two systems is key to developing alternative architectures that may address this problem.

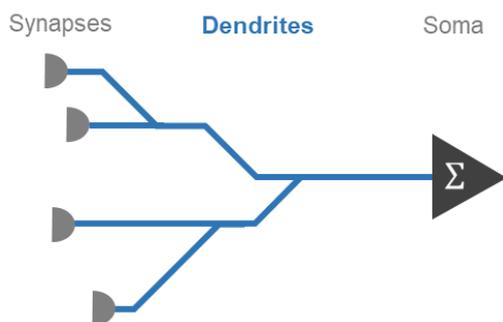

*Fig 1. Illustration of the location of dendrites within biological dendrites. Dendrites form a tree like structure, connecting synapses to the soma of the neuron.*

A significant difference between deep neural network models and biological neural networks is in the interconnects connecting synapses to the soma of the neuron. For the rest of this paper we will refer to the soma of the neuron as simply the neuron; as is convention in this field. In neural networks these are ideal connections, transporting data instantaneously from synapse to neuron with no distortion. But in biology this is not the case. The dendrites connecting synapses to their neuron form complex tree-like structures whose branches delay, attenuate, and transform action potentials that are sent along them.

One example of dendritic computation is in the task of locating the direction of an incoming sound. [2] The delays introduced by the dendritic trees of the auditory nerves have been found to compensate for the difference in the time of arrival of the sound at the left and right ears. This difference between the time of the sound arriving at the left ear, in turn inducing activity within the auditory nerves, and the time of the sound arriving at the right ear encodes the angle from which the incoming sound originated from. For example, if the sound is directly in front, then the activities of the left and right auditory nerves occur simultaneously. If the source of the sound were to progressively shift to the left, the activity from the right ear would occur later and later. To determine the angle from which the sound is coming from, the auditory nerves must therefore decode the time difference between these two signals.

The dendritic tree assists in this computation by forming branches each dedicated to detecting particular time differences. A single branch is tuned to detect a particular time difference by inducing a complementary delay to the signal from the left or right ear that brings the two activities back in line with each other. In doing so the amplitude of the total activity is increased, causing the attached neuron to fire. In contrast, if the time delay introduced by that specific branch does not match the time difference between the left and right activities then the amplitude of the total activity is not sufficient to cause the neuron to fire.

An array of these neurons each with a dendritic tree tuned to a particular time difference can therefore be used to decode the angle from which the sound originated, producing a single layer neural network with each output neuron corresponding to a particular range of angles. It has been suggested that with this architecture biological neurons can detect timing intervals with a resolution of 10-100$\mu s$ [3]–[6].

Dendrites have been replicated in the past. For example, one approach involves designing circuits which replicate models used by neuroscientists. For example, the linear cable model describes the basic electronic properties of the dendrite. It considers the dendrite as a conductive channel surrounded by an insulator layer. The insulating layer introduces a capacitance leading to leakage currents from the dendrite's core to the external environment for time varying signals. There is also an additional leakage term with equilibrates the membrane potential to a resting value.

Attempts to replicate this model use the silicon channel within a MOSFET to imitate the leakage current which is then placed in parallel with a capacitor. [7] This circuit replicates a single section of the dendrite and would be chained together to form longer branches of dendrites. This implementation has been shown to replicate the response of the linear cable theory model under small voltage biases, however for larger voltages the circuit's behaviour diverges from the model.

This design of dendrite has been applied to the problem of *word spotting* (detecting certain words within recorded audio) as part of a winner takes all network. [8]

While the above attempts to model the linear cable theory, other studies attempt to go deeper and replicate concentrations of specific ions within the dendrite. For example, the effect of calcium and sodium concentrations have be replicated in analogue circuits to generate the potential within the dendrite [9]. This results in a more complex circuit implementation than but could potentially lead to richer dynamics.

Alternatively, the dendrite may not be directly implemented but certain properties of the dendrite are replicated. For example, the attenuation and delay properties of dendrites have be reproduced with non-volatile memory crossbar circuits so as to obtain spike time dependent plasticity (STDP) behaviour [10]; where others use a silicon channel to implement the "add" and "subtract" properties. [11]

Irrespective of how we choose to replicate dendrites, there remains the question of how dendrites should be used, what impact they have on neural network performance and what properties of the dendrites are important for computation. These questions are important to answer because they inevitably influence our approach to constructing dendrites. The answer to how dendrites should be used, defines which properties should be replicated; while identifying the desired properties influences which technologies should be used to build dendrites.

In this paper, we focus on these fundamental questions, while remaining agnostic to the technology chosen to construct dendrites. We present a generic framework which can be translated across domains whether it be electronic, photonic or another medium. We first present a compilation of properties observed within biological dendrites and then identify applications of dendrites to demonstrate their potential impact. These example applications are not exhaustive, but we hope will inspire new uses of dendrites.

## 2 The Framework

### 2.1 Dendritic Properties

Dendrites have a range of properties, both passive and active. These properties can lead to carrying out computations within dendritic trees, or they enable a higher bandwidth within the channel. We have identified a subset of properties that have been observed in a variety of biological dendrites which we describe below. For each property we identify key descriptors which quantify that property and can be used to compare different technologies.

When referring to biological behaviours, we choose to focus on the effect of a behaviour and not on the biochemistry that causes it. Our justification for this decision is our wish to remain agnostic to the implementation of artificial dendrites. Understanding the biochemistry deriving these behaviours may indeed help inspire novel methods of building dendrites, but it is not necessary if only specific behaviours are of interest.

*A. Delay Lines*

Like any communication channel, dendrites inherently induce a time delay in the transmission of an input. Unlike a communication channel, this delay is desirable. The latency induced as the action potential travels along the dendrites can have computational applications [12]. For example, with synaptic inputs located at different points along the dendrite, action potentials from those closest to the neuron's soma arrive earlier than inputs situated further along the dendrite. The difference in arrival times of action potentials essentially labels them based on the location of the dendrite from which it was generated. There is evidence that this is used in biology to identify the location of noises in baby chicks. The ability to induce delays within neural networks may play a role in analysing time encoded data – an example of this is documented the following section.

The key descriptor of the delay line is how much delay is introduced per unit length. Depending on its implementation this could take a range of values and have a limitation based on the speed of propagation of the signal in the medium in question. Identifying this characteristic will also indicate the physical sizes required of the dendrites. Naturally faster mediums will require longer dendrites, thereby increasing circuit size. Therefore, to reduce circuit footprint a larger delay per unit length is desirable.

*B. Frequency Dependent Broadening & Low Pass Filtering*

Until now we have considered the delay induced by a dendrite as constant or at least assumed we were considering the propagation of an entire action potential. However, an action potential is a complex function made up of a combination of frequency components, each of which propagate through the dendrite at different speeds. This results in the different frequency components of the action potential experiencing different delays which in turn causes a broadening of the action potential. The dendrite is essentially acting as a low pass filter, broadening and attenuating the action potential.

However, the broadening of action potentials is not fixed. Instead, the extent to which the action potential is broadened can depend on the frequency of the spike train and previous action potentials. For example, in rat hippocampal pyramidal neurons it was observed that for higher frequency spike trains, each action potential is broadened to a greater extent [13]. The behaviour is therefore more complex than a simple low pass filter.

We can represent this behaviour as a low pass filter with a tuneable cut off frequency. For higher frequency spike trains the cut-off frequency reduces, causing the action potential to broaden to a greater extent. We should note from Kim et al. [13] that the increase in AP broadening for higher spike trains is not instantaneous. Instead, it takes multiple spikes of the higher frequency action potentials for the change in broadening to occur. There is therefore some form of integration occurring to cause this effect. We could represent this with a state variable which defines the cut-off frequency and updates over time.

### C. Variable Propagation Speed

The delays for a given frequency component introduced by the dendrite have so far been considered constant; we now take this concept further and introduce the ability to vary propagation speed within a section of the dendrite based on some input such as synaptic activity. This results in a delay which is dependent upon some input activity. At this point the complexity of the dendritic channel is increasing beyond a fixed structure to something responsive to its environment.

The action potential velocity can be anisotropic, with the dendrite modulating the velocity according to local synaptic inputs. [14] There is also evidence to suggest the speed is not only dependent on synaptic activity but the neuron's firing threshold as well. For example, it has shown experimentally that the propagation speed of an injected action potential can be modulated by adjusting the neuron firing threshold via an external electric field [15]. This exploited a technique where an external electric field is used to modulate the neuron's threshold [16].

There are therefore multiple mechanisms by which the propagation speed of the action potential can be modulated, potentially enabling more responsive computations within the dendrites. This may be an important behaviour to replicate as it allows a dynamic control of the dendrites properties that can respond to inputs. For example, modulating the action potential velocity is akin to the phase/frequency modulation employed in communication systems. There is therefore the potential to encode multiple synaptic inputs onto a single spike train.

### D. Amplification

Evidence of amplification within the dendrites has been observed in a number of samples such as the spinal motor neurons of cats [17], within the dendrites of neocortical neurons [18] and in rat hippocampal dendrites [19]. It is argued that amplification is crucial to transmitting synaptic inputs along longer dendrites such as the motor neurons [12].

The amplification is also found to vary across the dendrite. In simulations of dendrites, the amplification throughout the dendrite was found to be strongest within the trunk of the dendrite ~24 fold and weakened at the outer branches ~1.7 fold. [19] This inversely correlated with the impedance of the dendritic branch, which varied from $54 M\Omega$ within the trunk to $1200 M\Omega$ at the outer branches.

Implementing gains within artificial dendrites will be crucial if we are to construct larger dendritic trees. Without this, action potentials will be attenuated as they propagate through the network eventually becoming negligible, this places an upper limit on network size. Additionally, the presence of gain suggests the possibility of a segment of dendrite acting similar to the soma of a neuron. If the gain is sufficient, the segment could resemble a firing event.

*E. Integration*

The integration within dendrites is complex. The linearity of the integration depends on the timing of the action potentials, their amplitudes, and their spatial separation on the dendritic tree. In general, three different modes of integration are observed: sublinear, linear and supralinear [20].

To demonstrate the amplitude dependence of dendritic integration, previous studies observed the integration within hippocampal CA3 pyramidal neurons and found it to be linear for all action potentials with an amplitude <5mV. [21] In contrast, the linearity of action potentials with larger amplitudes depended on their timing and amplitudes. Sublinear integration was observed for larger amplitudes but only if the two inputs arrived at approximately the same time.

Although a variety of different integrations have been observed, researchers have observed some cells achieving an overall linear integration by exploiting sub and supralinear integration to overcome spatial effects such as attenuation along the dendrite [22].

The role of these different modes of integration is still debated. It may be that only linear summation is required to improve transmission of actin potentials, or it could be that the type of integration depends on the type of neuron the dendrites belong to. Without knowing the answer to this question, we should as designers attempt to replicate each of the integrations observed in biological neurons. This will give the designer a greater freedom when designing dendritic circuits and may also help in developing systems which can counteract nonideal transformations of the signal such as attenuation.

*F. Compartmentalisation*

Dendrite branches can be compartmentalised to varying degrees [23]. Compartmentalisation causes synapses to be independent of synapses within different compartments. This allows dendrites to selectively couple neighbouring synapses.

The definition of compartmentalization within the dendrite can vary depending on the study. It is sometimes referring to how electronically isolated one section of the dendrite is to another; the spines of a dendrite being a good example of this [24]. But it is also used to describe separate computing blocks, where a section of the dendrite exhibits plasticity while the other doesn't [25].

Despite these two being different explanations, we would argue the latter is the result of the former. Learning could not be compartmentalisation if there wasn't some form of electrical compartmentalisation. For example, the differences in plasticity responses in separate regions of the dendrite observed in [25] are explained via differences in calcium ion concentrations which is a form of electronic compartmentalisation.

Compartmentalisation occurs for a number of reasons, one of these being the dynamics of diffusion. The diffusion of ions spreading through the dendrite results in their concentration dropping as we move further from the injection site. Eventually this concentration becomes negligible and no longer affects the properties of the dendrite in this region [26], [27]. We can consider these sections compartmentalised from each other.

There are other causes of compartmentalisation such as the physical structure/dimensions of the section of dendrite. The spines are a good example of this. The constriction where the spines attach to the dendrite are on the order of approximately 140nm in diameter, slowing diffusion of ions between the spine and dendrite branch [28]. This also holds for the diameter of dendritic branches; thinner branches reduce the speed at which ions diffuse further isolating neighbouring sections of the dendrite [27].

The fact that compartmentalisation in biology occurs because of the diffusive properties of the ions and neurotransmitters in the dendrite suggest that the same section of dendrite could exhibit different compartmentalisation properties based on the ionic species. This is one potential advantage that dendrites have over electrons travelling through a wire.

### G. Multiplexing

How action potentials are multiplexed within neurons and dendrites is a debated topic. It is not clear whether the dendrites actively multiplex action potentials of different ionic species. Instead, superposition is more evident. Because of this multiplexing relies on frequency division multiplexing[29]–[31], time division multiplexing or more complex encoding schemes [32].

But there are systems where multiplexing of signals can be achieved within the same channel while exhibiting minimal crosstalk. For example, in optical communications wavelength division multiplexing is used to combine multiple data streams into a single fiber. Introducing multiplexing could help to avoid the challenge of routing dense networks of dendrites. While biological dendrites can achieve a connectivity, density and complexity we will struggle to replicate in solid state electronics, we could alleviate this issue by multiplexing different dendritic paths into a single channel.

## 3 Dendritic Computation Examples

Having identified the computational properties of dendrites and a biological example of dendritic computation, we need to consider where it could prove useful in neuromorphic systems. Currently neuromorphic circuits focus on trainable synapses that allow for generalized computing. This is a flexible approach that can be applied to a number of computational problems. In contrast, a dendritic approach does not appear to have the same flexibility. We have seen from the biological example of sound localisation that the structure of the dendritic tree, i.e. the length of its branches, plays a crucial role in the computation. Assuming this structure is fabricated in a particular form to solve a specific computational problem, the system is essentially hardcoded. It is not simple to adjust the structure of the dendrites after fabrication. For this reason, dendritic computation is more likely to suit computational tasks that are fixed. Pre-processing stages of input signals are an example of such tasks.

### 3.1 Pre-processing: Data conversion

Taking the biological example where dendrites aid in the localisation of a sound, the dendritic tree is essentially carrying out a decoding or transformation of the data to later be analysed by a neural network as illustrated in Fig 2B. Initially the information of interest (the direction of the sound) is encoded by the difference in the time of arrival of the sound between the left and right ears. The dendritic tree transforms this representation of the data into an array of neurons each of which correspond to a particular angle from which the sound arrived from – a spatial encoding. This encoding of the data may be a more efficient format for classification by a neural network.

To demonstrate the effectiveness of dendrites in this role, we have replicated this same sound localisation problem, comparing a neural network with and without dendritic pre-processing. The task of the neural network is to classify the timing separation of two pulses. We have simulated the accuracy for networks of different sizes to gauge whether dendrites can enable a network to obtain higher accuracies at smaller network sizes in comparison to a network without dendrites.

The dataset for the task is a collection of artificial audio pulses. Two copies of a 440Hz sinewave pulse are generated with a width of 10ms. One pulse is artificially delayed simulating the pulses arriving at each ear at slightly different times. Noise is added to each of these signals to resemble a real-world scenario, this consists of random additive white noise and jitter noise to disturb the timings of the signals. The dataset consists of three delays: 0, 5 and 10ms.

The neural network, illustrated in Fig 3A, consists of an input Bidirectional Long Short Term Memory (BiLSTM) layer [33] and one fully connected output layer. The BiLSTM layer introduces short and long term memory to the network and enables the classification of timeseries data. This layer has 2 inputs while the number of hidden nodes is varied from 1 to 10 to assess the impact of network size. The BiLSTM layer feeds into the fully connected output layer with 3 outputs, each corresponding to one of the 3 delays the network must classify.

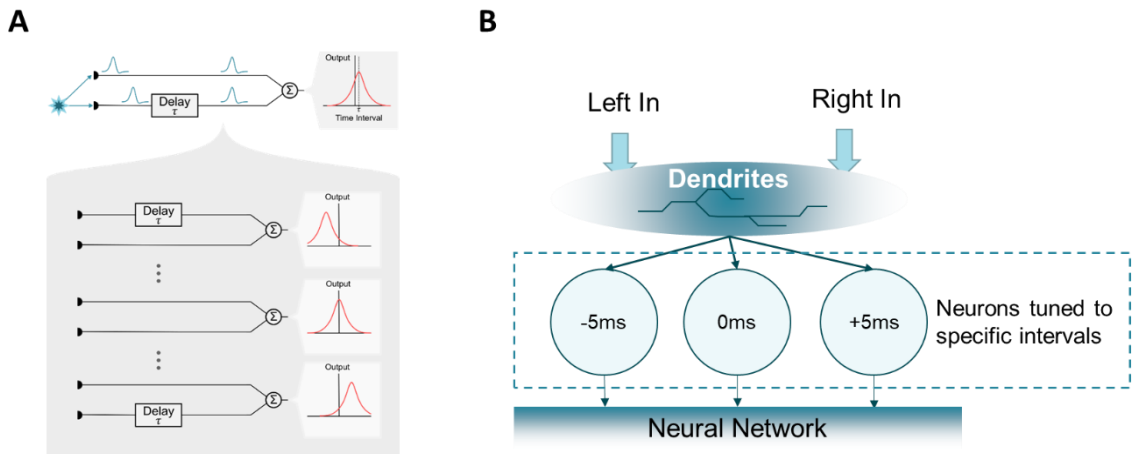

*Fig 2. Sound localisation with dendritic pre-processing. A) Dendrites in the auditory nerves assist neurons to detect phase differences from the left and right ears by inducing delays to synaptic inputs. By varying the delay of one channel with respect to the other, synaptic inputs will overlap when their phase corresponds to the delay induced by the dendrites. This results in a neuron which is most active to inputs separated by this specific delay. This sensitivity is tuned by varying the dendritic delays. B) An illustration of dendrites used as pre-processing elements. The dendrites convert time encoded inputs, from the left and right channels, into a spatial encoding where each neuron corresponds to a specific timing interval.*

The weights of the fully connected layer are trained using the Adam optimiser [34], a variant of stochastic gradient descent. We are primarily interested in the rate of learning for a given network size. Because of this, we limit the training to 50 epochs and validate the network's performance regardless of the accuracy reached during training. This allows us to compare the rate of learning for a network with and without dendrites.

Dendrites are placed at the input of the network. Each dendrite is designed to induce a delay to the inputs. We set one of the two input channels as the reference which is passed to the LSTM network with no delay. The second channel is connected to two dendrites, one induces no delay while the second induces a 10ms delay corresponding to the largest delay within the datset. With the dendrite layer now outputting three signals: the first channel with no delay and two copies of the second channel with increasing delays; the input to the LSTM layer is expanded from two to three inputs. The fully connected output layer is not changed.

We have trained both the plain BiLSTM network and the dendrite augmented BiLSTM network to evaluate their accuracy at classifying the pulse timing separations from our generated dataset. Their performances are plotted in Fig 3B for an increasing size of BiLSTM network. Each size of network was trained for a maximum of 50 epochs on a training dataset with 800 samples. The network's accuracy was validated on a separate test dataset also of 800 samples. Due to the stochastic nature of training, we repeated this process 40 times for each network size and have plotted the mean accuracies of these trials in Fig 3B.

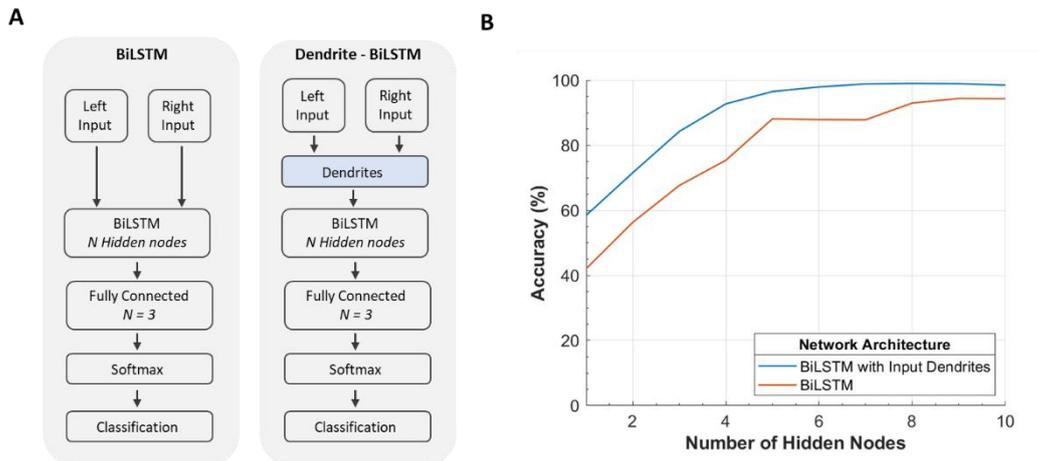

*Fig 3. Impact of an input dendrite layer on a BiLSTM network's performance. **A**) The architecture of a standard BiLSTM neural network and dendrite augmented BiLSTM network. Dendrites pre-process inputs from the left and right channels by introducing a 10ms delay to the right channel. **B**) The performances of the two networks are plotted for an increasing number of hidden nodes in the BiLSTM layer. Each sample point is the mean average of 40 trials with each trial limited to 50 epochs of training iterations.*

Both networks improve in accuracy with increasing network size, as would be expected. Equally, if both networks were allowed to train for a sufficient amount of time they could achieve similar performances. However, our focus is on reducing training times and network size. The network with input dendrites achieves a higher accuracy with a smaller LSTM network. It can be seen that for smaller networks (number of hidden nodes < 4), the dendrite augmented network achieves the same accuracy as a BiLSTM network with approximately one additional hidden node. For larger networks, the dendrite augmented network achieves the same performance as a network double its size. Dendrites have therefore enabled a reduction in network size.

The transformation of input data is one example of a pre-processing task that could be of use, but there are others such as compression. Biological systems take input from a large array of sensory cells however large numbers of input/output pins present a problem for neuromorphic processors. Dendritic trees could play a role in compressing data from multiple sensors into a single spike train: filtering unwanted data, transforming and combining multiple data streams into one via modulation of an action potential's velocity.

Finally, we have so far only considered the effect of dendrites at the input to a network. However, dendrites have potential within the inner layers as well. Especially, when we consider the integration and gain properties of a segment of dendrite. If the gain of a section of dendrite is sufficient enough, its behaviour could resemble that of a neuron by producing a large action potential from smaller input currents. This suggests we could

# 4  Conclusion

Dendrites are sophisticated communication channels with the ability to carry out computations within the branches of their trees. By omitting dendrites, we risk omitting a significant component of biological neural networks which could contribute to building more compact and efficient hardware neural networks.

We have investigated this question by remaining agnostic to any particular technology or method of implementing dendrites. We first identified key behaviours of biological dendrites from the relevant literature and presented these a framework for replicating dendrites. Not all these behaviours will be necessary, instead it depends on what context the dendrites are being used within.

To demonstrate the impact of dendrites, we replicated a biological example of dendritic computation. This involved using dendrites as a pre-processing layer to an LSTM neural network tasked with classifying the timing separation between pulses from two inputs. This replicates the role of dendrites within mammalian auditory nerves which aide neurons in locating sounds based on the phase difference between inputs from the left and right ears.

We found via simulation that an input dendritic layer can reduce the size of the LSTM network without a reduction in performance. Such reductions in network size, along with the associated reduction of training times, suggests dendrites could help to address the issue of power efficient neural networks on edge devices.

However, there remain a number of open questions on the use of dendrites, such as where dendrites have the most potential for impact on network efficiency. We have presented one example of dendrites at the input of a network, but it is equally possible to include dendrites within the inner layers. There is also the question of generalisability of a network containing dendrites. The example of pre-processing relied on knowledge of the problem being solved, this may be acceptable in an application which remains relatively constant, however if designing a chip for general learning problems, this may prove an issue.

Despite these uncertainties, dendrites clearly play an important role in biology and could have an impact on bio-inspired computing. We have presented a framework to guide the fabrication of artificial dendrites and an example application justifying their inclusion in neuromorphic circuits.